# Leveraging large language models for structured information extraction from pathology reports

Jeya Balaji Balasubramanian[a,*,†], Daniel Adams[b,c,†], Ioannis Roxanis[d], Amy Berrington de Gonzalez[b,c], Penny Coulson[c], Jonas S. Almeida[a], Montserrat García-Closas[b,c]

[a] Division of Cancer Epidemiology and Genetics, National Cancer Institute, Rockville, MD, USA.
[b] The Cancer Epidemiology and Prevention Research Unit, The Institute of Cancer Research, London and Imperial College London, England.
[c] Division of Genetics and Epidemiology, The Institute of Cancer Research, London, England.
[d] The Breast Cancer Now Toby Robins Research Centre, The Institute of Cancer Research, London, England.

[*] Corresponding author.
[†] Authors contributed equally.

## Authors

Dr. Jeya Balaji Balasubramanian
Division of Cancer Epidemiology and Genetics, National Cancer Institute, Rockville, MD, USA.
9609 Medical Center Dr, NCI Shady Grove, Room 7E554, Rockville, MD, USA 20850.
jeya.balasubramanian@nih.gov
+1 240-276-6385

Dr. Daniel Adams
The Cancer Epidemiology and Prevention Research Unit, The Institute of Cancer Research, London and Imperial College London, England.
Division of Genetics and Epidemiology, The Institute of Cancer Research, London, England.
daniel.adams@icr.ac.uk

Dr. Ioannis Roxanis
The Breast Cancer Now Toby Robins Research Centre, The Institute of Cancer Research, London, England.
ioannis.roxanis@icr.ac.uk

Prof. Amy Berrington de Gonzalez

The Cancer Epidemiology and Prevention Research Unit, The Institute of Cancer Research, London and Imperial College London, England.
Division of Genetics and Epidemiology, The Institute of Cancer Research, London, England.
amy.berrington@icr.ac.uk

Penny Coulson
Division of Genetics and Epidemiology, The Institute of Cancer Research, London, England.
Penny.Coulson@icr.ac.uk

Dr. Jonas S. Almeida
Division of Cancer Epidemiology and Genetics, National Cancer Institute, Rockville, MD, USA.
jonas.dealmeida@nih.gov

Prof. Montserrat García-Closas
The Cancer Epidemiology and Prevention Research Unit, The Institute of Cancer Research, London and Imperial College London, England.
Division of Genetics and Epidemiology, The Institute of Cancer Research, London, England.
montse.garcia-closas01@icr.ac.uk

# Abstract

## Background

Structured information extraction from unstructured histopathology reports facilitates data accessibility for clinical research. Manual extraction by experts is time-consuming and expensive, limiting scalability. Large language models (LLMs) offer efficient automated extraction through zero-shot prompting, requiring only natural language instructions without labeled data or training. We evaluate LLMs' accuracy in extracting structured information from breast cancer histopathology reports, compared to manual extraction by a trained human annotator.

## Methods

We developed the Medical Report Information Extractor, a web application leveraging LLMs for automated extraction. We developed a gold standard extraction dataset to evaluate the human annotator alongside five LLMs including GPT-4o, a leading proprietary model, and the Llama 3 model family, which allows self-hosting for data privacy. Our assessment involved 111 histopathology reports from the Breast Cancer Now (BCN) Generations Study, extracting 51 pathology features specified in the study's data dictionary.

## Results

Evaluation against the gold standard dataset showed that both Llama 3.1 405B (94.7% accuracy) and GPT-4o (96.1%) achieved extraction accuracy comparable to the human annotator (95.4%; p = 0.146 and p = 0.106, respectively). While Llama 3.1 70B (91.6%) performed below human accuracy (p <0.001), its reduced computational requirements make it a viable option for self-hosting.

## Conclusion

We developed an open-source tool for structured information extraction that can be customized by non-programmers using natural language. Its modular design enables reuse for various extraction tasks, producing standardized, structured data from unstructured text reports to facilitate analytics through improved accessibility and interoperability.

# 1. Introduction

Structured information extraction [1] from unstructured histopathology reports facilitates data accessibility and analytics for clinical research. Manual structured information extraction is time-consuming and expensive, limiting its scalability for large datasets. Furthermore, it necessitates domain expertise and is susceptible to human error. Natural Language Processing (NLP), a subdomain of Artificial Intelligence (AI), offers methods for automated and scalable structured information extraction from large volumes of unstructured text reports. [2] However, conventional NLP approaches often require task-specific model training with manually labeled data, which can be challenging to obtain, and the resulting models typically have limited generalization capability to unseen scenarios compared to more recent approaches. [3,4] To address these challenges, efficient NLP solutions are desirable that can reduce the reliance on labeled data while offering improved generalizability.

Foundation models represent an emerging new paradigm in developing AI-based solutions. [5] These are large-scale machine learning models pre-trained on extensive and diverse datasets to provide a generalist model which can be adapted for a wide range of specialized tasks. While some foundation models can support multimodal inputs—meaning they can process multiple types of data such as text, images, and audio simultaneously—our study focuses solely on text-based models. Large language models (LLMs), with state-of-the-art examples like OpenAI's GPT-4o [6,7] and Meta's Llama 3, [8] are types of foundation models capable of processing and generating natural language texts. GPT-4o and Llama 3 have also been tuned to follow instructions in line with human expectations.

Two prominent techniques for adapting LLMs for specialized tasks are prompting and fine-tuning. [5] Prompting involves providing the LLM with task-specific instructions and examples in natural language. Fine-tuning requires additional model training on task-specific data. Prompting is simpler as it needs little to no data and no model training. If prompting yields inadequate performance, fine-tuning can be considered. Zero-shot prompting is a simple strategy, where only the task instructions are provided to the LLM, without any examples of the task being solved. [9] This approach is particularly efficient since it requires no labeled data or training while potentially offering better generalization than conventional NLP approaches. [3,4]

We evaluate the accuracy of an LLM-based solution for automated structured information extraction compared to a manual approach by domain experts. Specifically, we assess state-of-the-art LLMs adapted through zero-shot prompting for the specialized task of structured information extraction from breast cancer histopathology reports. Our assessment involves 111 anonymized histopathology reports from the Breast Cancer Now (BCN) Generations Study in the United Kingdom. From these reports, we extract 51 pathology features, as specified by the BCN Generations Study histopathology data dictionary. Additionally, we standardize the extracted data using SNOMED CT (Systematized Nomenclature of Medicine - Clinical Terms) to facilitate collaborative research through either interoperable data sharing [10] or decentralized analytics. [11] Following the Findable, Accessible, Interoperable, and Reusable (FAIR) principles for epidemiologic research, [12] we store the standardized data in a non-proprietary, machine-readable format.

Recent works have explored zero-shot prompting of LLMs for structured information extraction from pathology reports. Choi et al. (2023) [13] evaluated OpenAI's GPT-3.5 in extracting 17 features from 340 breast cancer pathology reports from Seoul National University Hospital. Truhn et al. (2024) [14] evaluated GPT-4's extraction capabilities across two domains: 5 features from TCGA colorectal cancer reports and 11 features from University College London Hospitals' diffuse glioma reports (100 reports each). Huang et al. (2024) [15] conducted a larger-scale but narrower extraction, using GPT-3.5 to identify 4 features from 852 TCGA lung cancer reports and 2 features from 191 osteosarcoma reports at the University of Texas Southwestern Medical Center. Our work differs from previous studies in several key aspects: we extract 51 distinct features from 111 breast cancer reports, significantly expanding the scope of information extraction; we evaluate newer models, including GPT-4o, and the self-hostable Llama 3 model family for enhanced data privacy; we assess the models' ability to conform to rigid formatting constraints specified by the study's data dictionary without any post-processing of the LLM's output; and we provide an open-source, reusable tool that can be adapted for any generic structured extraction task.

The subsequent sections are outlined as follows. 'Materials and methods' (section 2) describes the breast cancer histopathology reports and the associated data dictionary from the BCN Generations Study, our approach for data standardization, and manual information extraction by experts. We introduce the Medical Report Information Extractor, our web application for automated information extraction leveraging LLMs. The section also describes our evaluation framework, including data preprocessing, the LLMs assessed, and our validation approach. 'Results' (section 3) presents the performance comparison between the manual and automated extraction processes. Finally, 'Discussion' (section 4) examines the results while reflecting on the performance-cost trade-offs, challenges and limitations of our approach, practical considerations when developing AI solutions with foundation models, and directions for future work.

# 2. Materials and methods

## 2.1 Generations Study histopathology data

### 2.1.1 Data description

The BCN Generations Study is a large-scale prospective cohort study initiated in 2003 in the United Kingdom. Focused primarily on investigating breast cancer etiology, the study has enrolled over 113,000 women aged 16 years and older. Participants provided questionnaire information and informed consent upon recruitment, including consent to access medical records. The study received approval from the South-East Multicentre Research Ethics Committee.

As part of the BCN Generations Study, histopathology reports were collected from participants who developed breast cancer within the follow-up period. For this analysis, the reports of 125 participants were randomly selected and made available as scanned images in Tagged Image File Format (TIFF). The initial dataset consisted of 133 breast cancer pathology reports dated between 2010 and 2020 collected from 125 participants. Participants with multiple reports were excluded due to uncertainty in matching the previously manually-extracted data with the corresponding reports. Reports containing multiple tumor specimens were also excluded, as it was unclear to which specimen the manually-extracted data referred to. After excluding 22 reports from 14 participants, the final dataset included 111 breast cancer pathology reports, each corresponding to a single participant and describing a single tumor specimen.

To ensure confidentiality, the pathology reports and associated data were fully anonymized through a combination of automated and manual de-identification processes, ensuring all personal identifiers were removed prior to processing by the LLMs. These anonymization methods are described in detail under the 'Data preprocessing' header within the 'Evaluation' subsection (section 2.4.1).

### 2.1.2 Data schema

The BCN Generations Study histopathology data dictionary comprises 51 features of interest. These features can be categorized into broad groups, including specimen characteristics, breast lesion type, morphological and architectural features of the lesions, tumor size measurements, surgical margin assessments, lymphovascular status, molecular scoring, tumor grading and staging, and TNM classification. For each feature, the dictionary provides comprehensive technical metadata, specifying the data type, format, measurement units, value range constraints, and categorical value codings. Additionally, the dictionary includes descriptive metadata in the form of human-readable descriptions of both the features and their corresponding values, facilitating interpretation of the dataset.

The BCN Generations Study histopathology data dictionary was originally in the comma-separated values (CSV) format. When this CSV-formatted metadata was input into LLMs for information extraction, we observed that the models frequently confused technical and descriptive metadata across different features, particularly when tasked with extracting a large number of features simultaneously. To address these limitations, we turned to JavaScript Object Notation (JSON) Schema, a specification for a JSON-based file format that serves a purpose similar to a data dictionary for JSON-formatted data. [16] JSON Schema allows for a structured representation of both technical and descriptive metadata, making it particularly well-suited for conveying complex information. We reformatted the BCN Generations Study histopathology data dictionary as JSON Schema. When the metadata was provided to LLMs as JSON Schema, it significantly improved their ability to correctly process the metadata information simultaneously for a large number of features.

## 2.1.3 Data standardization

To facilitate collaborative research, we standardize data extracted from pathology reports using SNOMED CT, a comprehensive clinical terminology system. We implement this standardization through Linked Data technology, which is a way to create links between structured data on the Web. [17] These links are established using Internationalized Resource Identifiers (IRIs), which are standardized identifiers that uniquely label concepts across the Web. When multiple datasets reference the same concepts using these IRIs, researchers can discover and integrate related datasets by following these conceptual links. We format the standardized extracted data using JSON-LD (JavaScript Object Notation for Linked Data), a JSON-based format designed for expressing Linked Data. [18]

JSON-LD allows us to specify a context file that serves as a semantic dictionary. It maps local feature and value names to corresponding standardized SNOMED CT terms using IRIs. For example, the BCN Generations Study histopathology data dictionary specifies a feature named "Side" to indicate the laterality of tissue specimens. We use the JSON-LD context file to map the ambiguous local feature name "Side" to the standardized SNOMED CT term "Specimen laterality" (ID: 384727002) through its corresponding SNOMED CT IRI. This mapping ensures that other researchers and systems can unambiguously interpret this feature in our dataset.

We mapped features and values in the BCN Generations Study histopathology data dictionary to standardized SNOMED CT terms using an automated semantic term matching algorithm that we developed. Other tools for semantic terminology mapping exist, such as Usagi, [19] an open-source tool developed by Observational Health Data Sciences and Informatics (OHDSI). A comprehensive evaluation of automated data standardization approaches remains a direction for future research. For the purposes of this study, we assume pre-existing mappings between local data dictionary terms and standardized vocabularies. Furthermore, our proposed solution treats data standardization as an optional step, allowing flexibility in implementation while focussing on the task of data extraction.

## 2.2 Manual structured information extraction

An expert human annotator manually extracted data for the 51 features specified in the data dictionary from the 111 pathology reports. This manually extracted data was de-identified for the purposes of evaluation. In contrast to LLMs, which require text-based inputs, the human annotator worked directly with the physical reports. LLM-based solutions, however, require preprocessing steps such as converting scanned reports into machine-readable texts for LLM processing, and de-identification to ensure data privacy. These steps, while necessary for LLM input, can introduce potential bottlenecks and impact their overall performance.

## 2.3 Medical Report Information Extractor

We developed a web application called "Medical Report Information Extractor", which leverages LLMs to automatically extract structured information from unstructured pathology reports. Figure 1 illustrates the architecture of this application.

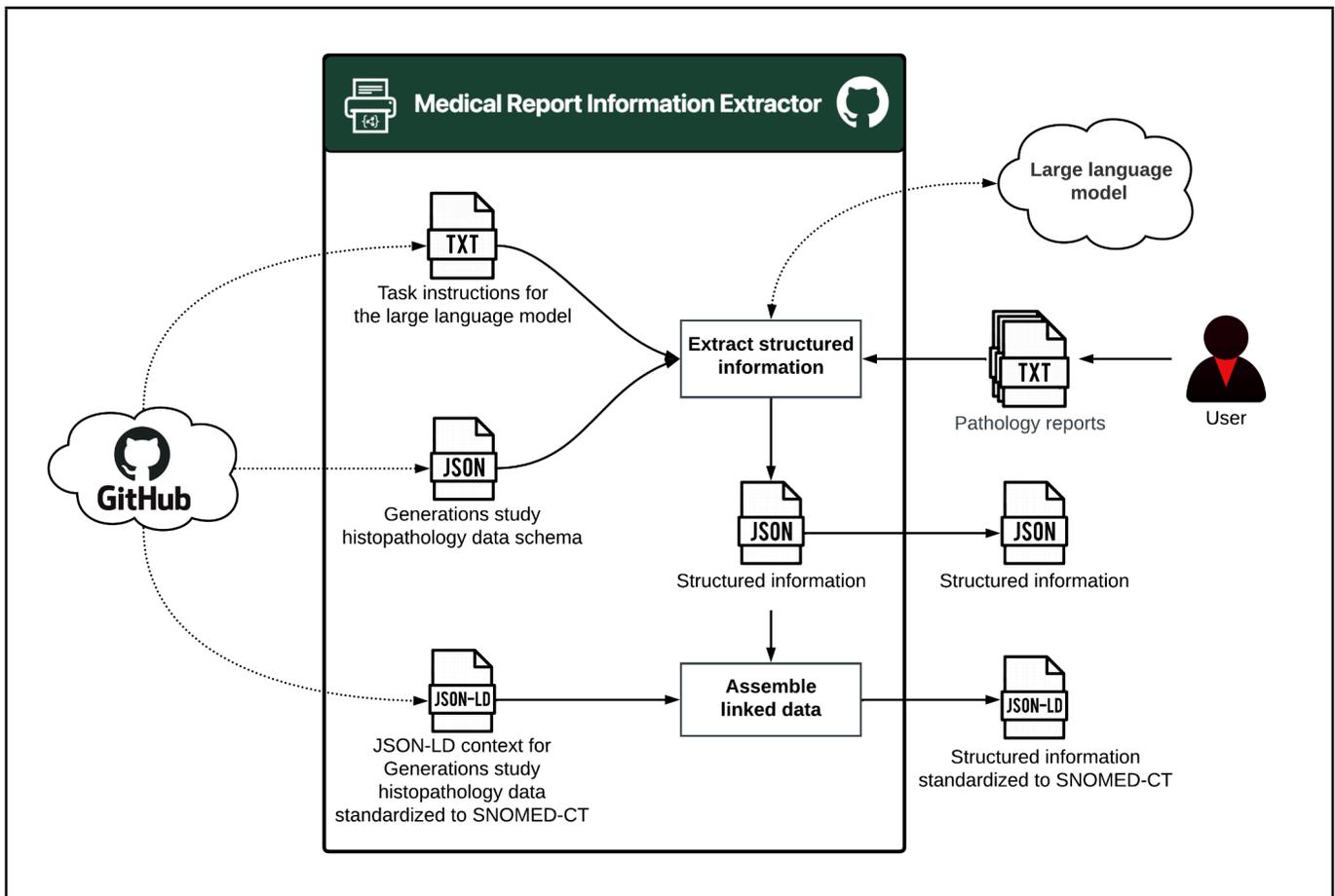

**Figure 1: Schematic of the Medical Report Information Extractor application architecture.** This web application leverages LLMs to automatically extract structured information from unstructured pathology reports. Users input plaintext pathology reports and specify an API endpoint serving an LLM of their choice. The application outputs the LLM-extracted structured information in JSON format, and a corresponding linked data in JSON-LD format, which standardizes the data to SNOMED CT. The application's behavior is customizable through three external application configuration files: 1) a text file containing the task instructions for the LLM, 2) a JSON Schema file specifying the structure and description of the information to extract according to the BCN Generations Study histopathology data dictionary, and 3) a JSON-LD context file that maps the feature and value names in the JSON Schema to standardized concepts in SNOMED CT.

The users input plaintext pathology reports, and an API endpoint serving an LLM of their choice. Specifically, the users provide a base URL and an API key to an endpoint compatible with OpenAI's API structure. The application first queries the 'models' endpoint to retrieve the list of models available to the user. The user then selects the LLM they wish to employ for the task. Subsequently, the application utilizes the selected LLM via the 'chat completions' endpoint to extract structured information from the pathology reports. The application returns the structured information in JSON format, and corresponding linked data in JSON-LD format that standardizes the extracted data.

This design allows users to work interchangeably with various LLMs, including those that can be self-hosted and run locally. Tools like vLLM enable users to self-host downloadable LLMs, such as Llama 3, on OpenAI-compatible API servers. [20] Alternatively, many LLMs are conveniently available off-the-shelf as fully managed services, hosted on platforms such as OpenAI, Anthropic Claude, Mistral AI, Amazon Bedrock, Azure AI Studio, and Google Cloud Platform's VertexAI. Our application design accommodates the use of LLMs hosted both locally and on such platforms. For managed services that do not provide an OpenAI-compatible API, an intermediary wrapper API can be created to mimic its structure.

The application's behavior can be customized through three external application configuration files:
1. A text file containing natural language task instructions for the LLM.
2. A JSON Schema file specifying the structure, format, and description of the information to extract from the pathology reports.
3. An optional JSON-LD context file mapping the data feature and value names in the JSON Schema to concepts in standardized vocabularies. When provided, the application outputs the data in JSON-LD format alongside the standard JSON, effectively transforming the structured data into standardized linked data.

To extract structured data specified by the BCN Generations Study histopathology data schema and to standardize it to SNOMED CT, we configure the application with the JSON Schema and the JSON-LD context file we described earlier in the 'Generations Study histopathology data' subsection (section 2.1.2 and 2.1.3, respectively). The task instructions in the text file were developed and the JSON Schema refined through the iterative prompt development process detailed in the 'Validation design' part of the 'Evaluation' subsection (section 2.4.3).

## 2.4 Evaluation

Our evaluation process aimed to assess, against the gold standard, the accuracy of five LLMs and a human as annotators in extracting structured information from pathology reports. This process is elaborated upon below and includes data preprocessing, prompt engineering, and gold standard data generation, followed by a statistical comparison of the annotators' extraction accuracy. Figure 2 provides a graphical overview of the evaluation process.

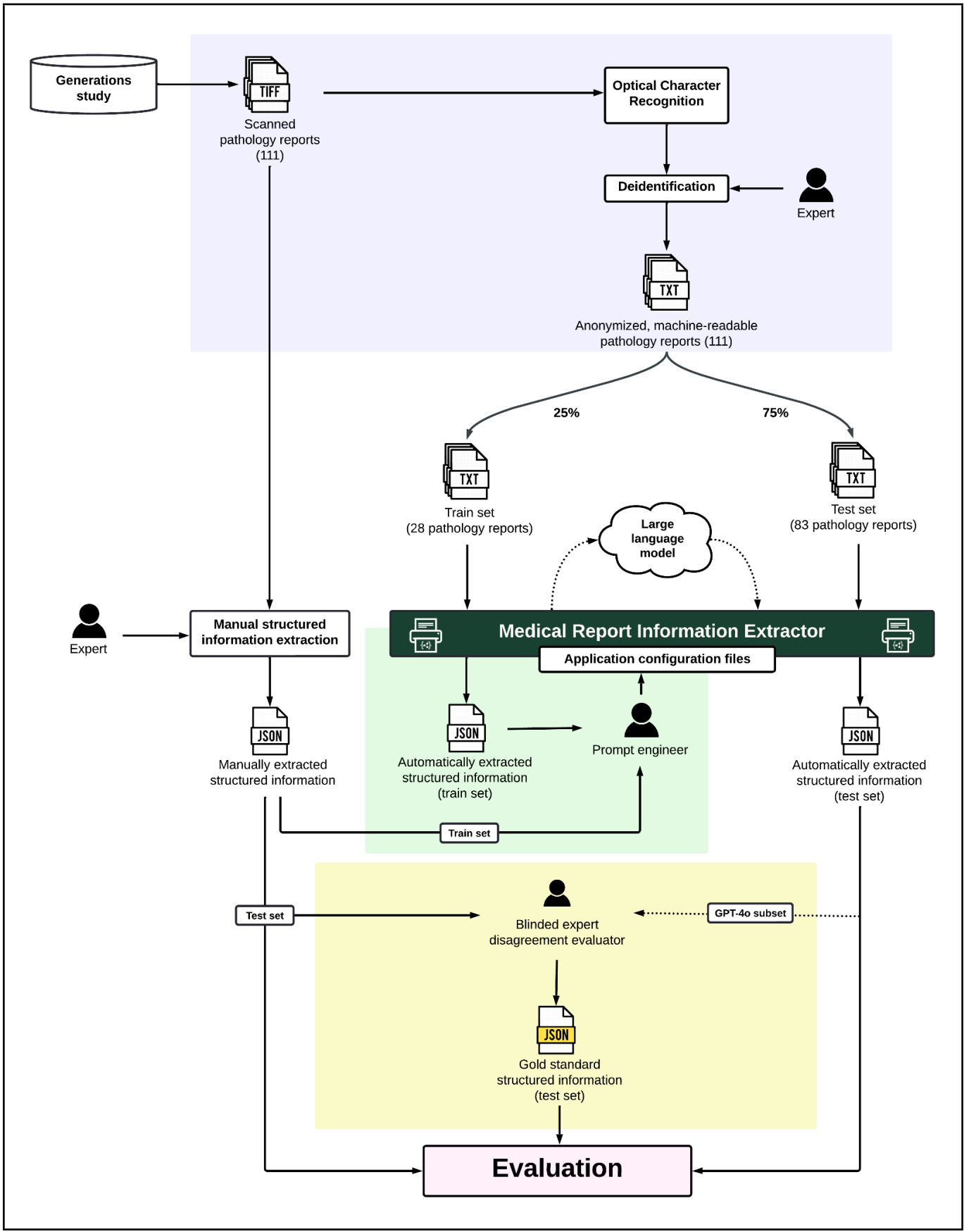

**Figure 2: Schematic of the structured information extraction and evaluation pipeline.** The pre-processing phase (blue) involves OCR and de-identification. The manual extractions performed by the human annotator also underwent de-identification. The training phase (green) consists of iterative prompt engineering using the training dataset. The gold standard dataset (yellow) is composed of GPT-4o and human agreements, with expert resolution of disagreements. The evaluation phase (red) compares all annotators to the gold standard on the test dataset.

### 2.4.1 Data preprocessing

Data preprocessing was performed to convert scanned pathology reports into de-identified, machine-readable text, to enable processing by LLMs. To this end, we developed a multi-step process involving optical character recognition (OCR), report content de-identification, and layout reassembly. To improve text recognition accuracy, the scanned report images were first converted to grayscale, corrected for skew, and binarized using local thresholding to improve contrast. Pseudonymized identifiers were assigned to the scanned reports, and the mapping between these identifiers and the original reports was not retained.

The DocTR [21] package, utilizing the DB-ResNet-50 text detection engine, [22] was applied to identify the bounding boxes of textual elements within the images. Once the textual elements were located, they were cropped from the reports, padded with whitespace, and processed through the Tesseract OCR engine, configured with a deep learning model for text recognition. Following this, the Presidio package was used for de-identification, [23] merging the results from two deep learning models pre-trained to detect personally identifiable information. [24,25] Regular expressions were applied to de-identify UK postcodes, as Presidio does not include them in its recognized entities.

To ensure that important spatial information, such as indentations and tabular elements, was not lost on conversion to machine-readable text, it was necessary to preserve the spatial arrangement of the original document. Although Tesseract's "whitespace-preserving" configuration was initially considered for this purpose, it introduced numerous textual artifacts. Instead, a custom solution was implemented, mapping the extracted text onto a plaintext grid based on their relative positions identified by DocTR, with the grid's resolution determined by the median character size. As a final step, the plaintext report underwent an additional round of de-identification by two independent human annotators. This process resulted in fully anonymized, plaintext pathology reports that were ready for input into the LLMs. Any coded link between anonymized reports/data and identifiers in the Generations database was destroyed to prevent re-identification.

The full preprocessing pipeline and code are available at https://github.com/rictoo/ocr-deidentifier.

### 2.4.2 Language models evaluated

GPT-4o is a leading proprietary LLM. Proprietary LLMs are typically hosted by external organizations, requiring users to send data to third-party servers. This raises significant data privacy concerns, especially with sensitive data like pathology reports, necessitating their de-identification. In contrast, Llama 3 is a leading publicly available LLM that allows users to self-host the model. Self-hosting LLMs ensures that confidential data remains within the organization's governance and helps comply with restrictive data privacy regulations, such as requirements for data de-identification. To understand the performance tradeoffs between these two categories of LLMs, we examine: Proprietary LLMs, represented by OpenAI's GPT-4o family, [6] and LLMs publicly available for download and self-hosting, represented by Meta's Llama 3 family. [8]

We evaluate two models from the GPT-4o family, accessed through the OpenAI API platform:
1. GPT-4o (API version gpt-4o-2024-05-13): Demonstrates state-of-the-art performance across traditional NLP benchmarks. [6]
2. GPT-4o mini (API version gpt-4o-mini-2024-07-18): A more cost-efficient alternative, offering competitive performance relative to similarly-priced models, albeit with reduced performance compared to GPT-4o. [26]

From the Llama 3 model family, we evaluate all three instruction-tuned models offered under version 3.1:
1. Llama 3.1 405B (version Meta-Llama-3.1-405B-Instruct): Demonstrates performance comparable to leading proprietary LLMs, including GPT-4o, across the key NLP benchmarks. [8]
2. Llama 3.1 70B (version Meta-Llama-3.1-70B-Instruct): A significantly smaller model compared to the 405B parameter version. If this model performs competitively for our task, it might be preferred over the 405B parameter model due to its reduced computational resource requirements for hosting and inference.
3. Llama 3.1 8B (version Meta-Llama-3.1-8B-Instruct): The smallest model in the Llama 3.1 family. Competitive performance from this model would be particularly advantageous due to its minimal resource requirements. Its portability allows it to be deployed entirely within the privacy of the user's web browser, as demonstrated by the WebLLM project. [27]

Amazon Bedrock is a fully managed service that provides API access to leading foundation models, including the Llama 3 model family, hosted on the Amazon Web Services (AWS) cloud platform. [28] Although self-hosting these models is possible, we chose to evaluate the Llama 3 models through Amazon Bedrock APIs for convenience in our current assessment.

Both OpenAI and Amazon Bedrock APIs ensure that user data is not utilized for model training and remains securely encrypted both at rest and during transit. [29,30]

### 2.4.3 Validation design

To evaluate the generalizability of an LLM adapted for a specialized task through prompting, we can employ a traditional cross-validation design. Our dataset of 111 breast cancer pathology reports was split into a training set (28 reports, 25%) and a test set (83 reports, 75%). The training set was used for iterative prompt development, guiding the LLM towards correctly extracting the desired information. Prompt optimization relied solely on training set performance, ensuring no test set information influenced the process. The independent test set was then used to validate both the LLMs' and the human annotator's information extraction performance. By developing prompts on a separate training set and evaluating on an independent test set, we can robustly assess how well the LLM generalizes to unseen pathology reports without bias due to training set contamination.

In this study, a prompt engineer manually performed iterative prompt development. The LLMs conducted automated structured information extraction via the Generations Pathology Information Extractor web application. In this application, prompts take the form of application configuration files, comprising a text file with task instructions for the LLM, and a JSON Schema file defining the structure and description of the information to extract, based on the BCN Generations Study histopathology data dictionary. The prompt engineer refined the JSON Schema's descriptive metadata to provide clarifications, specialized instructions, and warnings about potential edge cases. Additionally, the engineer grouped the 51 features to extract into 13 batches of logically related features, which improved performance.

## 2.4.4 Gold standard label generation

To evaluate the performance of both the human annotator and the LLMs, a gold standard dataset was generated. This process involved a physician who worked independently from the human annotator. The gold standard dataset was created by performing a conflict resolution between the automated structured information extraction by GPT-4o, a leading proprietary LLM, and the manual structured information extraction by the human annotator, a domain expert.

When GPT-4o and the human annotator agreed on a feature-value annotation, it was automatically accepted as the gold standard truth. The physician was only involved when GPT-4o and the human annotator disagreed. This focused approach helped optimally manage the physician's workload and minimize the potential for further human errors. While this approach is efficient, it could result in a small chance of incorporating erroneous annotations into the gold standard, as both GPT-4o and the human annotator may have been incorrect despite their agreement.

In cases of disagreement between GPT-4o and the human annotator, a conflict resolution process was initiated. A web application was developed to present the conflicting annotations to the physician in a randomized order. The physician was blinded to the source of the annotations and was provided with the corresponding anonymized, machine-readable pathology reports. Additionally, an independent instance of GPT-4o was made available to assist the physician in resolving conflicts. This assistant LLM was also blinded to the true source of the conflicting annotations.

After reviewing the pathology report, the physician provided an independent annotation, which could either agree with one of the existing annotations or introduce a new, third option. If a perceived OCR error resulting in the omission or distortion of information—such as inconsistent numerical values suggestive of digit insertions or replacements, erroneous redactions of relevant information, or illegible text—was identified by the physician, the gold standard feature-value annotation was assumed to be that of the human annotator. This methodology allowed for the creation of a robust gold standard dataset to evaluate the annotators.

### 2.4.5 Statistical analysis of annotator performance

To assess the performance differences among annotators, a statistical framework was employed that determined whether the extraction performance, in terms of alignment with the gold standard, significantly deviated from that of the human annotator. The approach was based on the assumption that the number of correctly extracted features, $k$, for each annotator within a given report followed a binomial distribution. The binomial distribution was parameterized by the total number of extractable features, $n$, and a latent variable $p$, which represented the probability of successful extraction, where success was defined as alignment with the gold standard.

$$k \sim \text{Binomial}(n, p) \tag{1}$$

This framework enabled a binomial logistic regression model to be fitted to the data where $p$, the probability of successful extraction, was modeled as a function of the annotator. Using this regression model, statistical comparisons were conducted using Wald tests to evaluate whether each annotator's performance significantly differed from that of the reference, chosen to be the human annotator. A p-value threshold of < 0.05 was used for statistical significance in this study.

$$\log\left(\frac{k}{n-k}\right) = \beta_0 + \beta_1 \cdot \mathbb{1}_{\text{GPT-4o}} + \beta_2 \cdot \mathbb{1}_{\text{GPT-4o mini}} + \beta_3 \cdot \mathbb{1}_{\text{Llama 3.1 405B}} + \beta_4 \cdot \mathbb{1}_{\text{Llama 3.1 70B}} + \beta_5 \cdot \mathbb{1}_{\text{Llama 3.1 8B}} \tag{2}$$

# 3. Results

A total of 4,233 extractions were performed across 83 reports and 51 features by each annotator. Of these extractions, 23 were flagged as suspected OCR errors, with the human annotator's extraction designated as the gold standard in those cases. Although multiple-specimen reports were initially identified for exclusion during dataset preparation, one extraction from a single report was removed during the analysis because the report was found to describe multiple specimens, leading to ambiguities in the feature being extracted. The performance of the annotators was evaluated on the remaining 4,232 extractions.

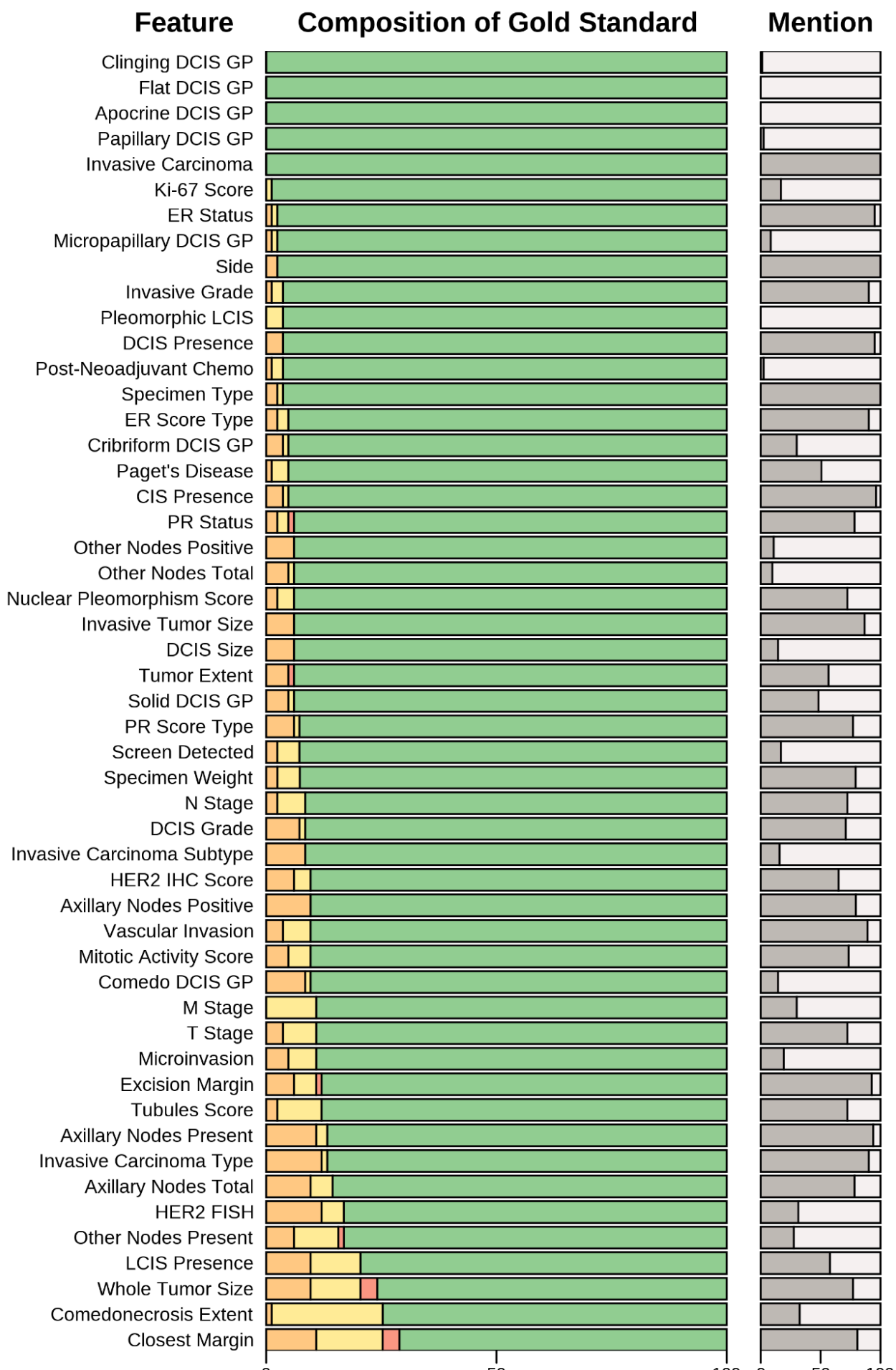

**Figure 3: Composition of the gold standard by feature (left) and the percentage of reports where each feature had a finding mentioned (right).** The green bars in the left bar plot represent the percentage of reports where GPT-4o and the human annotator agreed on the value of a feature. For cases of disagreement, a physician resolved the conflict: yellow indicates alignment with the human annotator, orange with GPT-4o, and red indicates a different value introduced by the physician. The bar plot on the right shows, in dark gray, the percentage of reports in which each feature was mentioned. Mention status was determined based on the gold standard: a feature within a report was considered unmentioned if 'no mention' was selected for those features where that option was available, and by the selection of 'negative' otherwise. Acronyms used in feature names are defined in Table S1 in the Supplement.

There was a 91.8% agreement (3,884 out of 4,232 extractions) between the human annotator and GPT-4o. Of the 348 disagreements reviewed by the independent evaluator, 52.9% (184 extractions) aligned with GPT-4o, 44.3% (154 extractions) aligned with the human annotator, and 2.9% (10 extractions) aligned with neither GPT-4o nor the human annotator.

Among the features with the highest number of disagreements between the human annotator and GPT-4o (Figure 3), multiple lymph node-related features were present. Disagreements also occurred with the comedo necrosis feature, where GPT-4o often assigned explicit absence rather than leaving the feature unspecified when reports made no mention of it. The human annotator did not make this error. The *Whole Tumor Size* and *Closest Margin* features also showed substantial disagreement between both annotators. The majority of gold standard disagreements with both annotators were also among these two features. We determined how often each feature was mentioned across reports (Figure 3), and no clear relationship was observed between the mention of a feature and the probability of agreement between the human annotator and GPT-4o across reports.

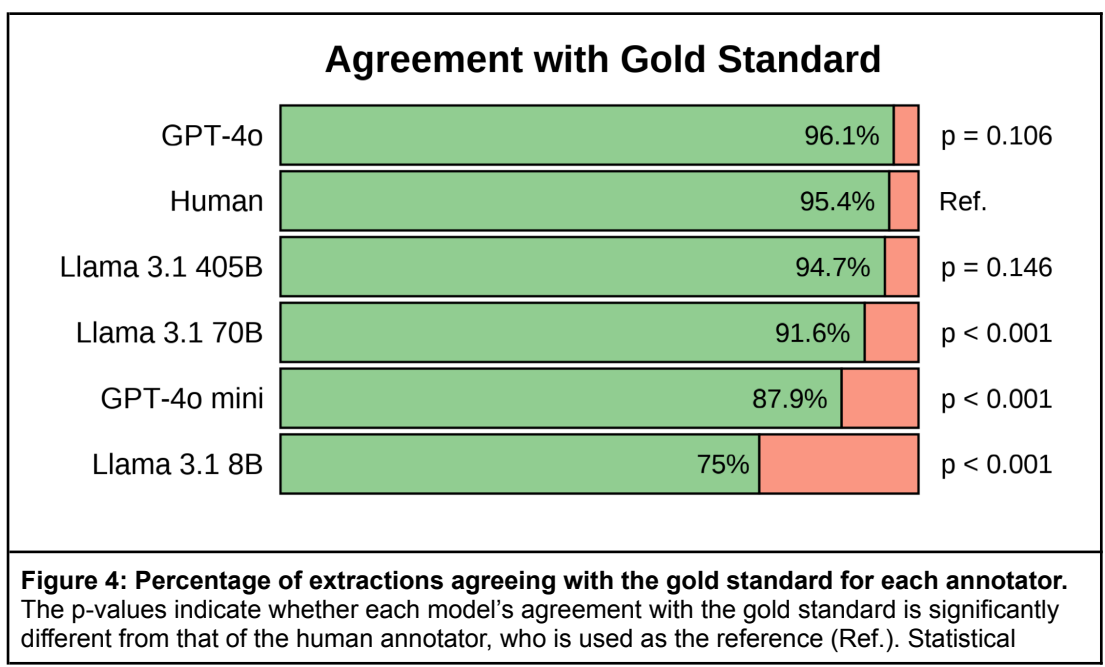

**Figure 4: Percentage of extractions agreeing with the gold standard for each annotator.** The p-values indicate whether each model's agreement with the gold standard is significantly different from that of the human annotator, who is used as the reference (Ref.). Statistical

> comparisons were conducted using Wald tests from a binomial logistic regression model.

Wald tests were conducted to evaluate statistically significant differences in annotator performance relative to the human annotator, in terms of alignment with the gold standard (Figure 4). Results revealed no significant difference in extraction accuracy between the Llama 3.1 405B and the human annotator. Among the Llama 3.1 model variants, we observed a clear trend where larger models demonstrated increased performance, aligning with the general expectation from increasing model size. Similar to Llama 3.1 405B, the GPT-4o model also did not differ statistically from the human annotator in terms of extraction accuracy. In contrast, the GPT-4o mini model outperformed only the smallest Llama 3.1 model.

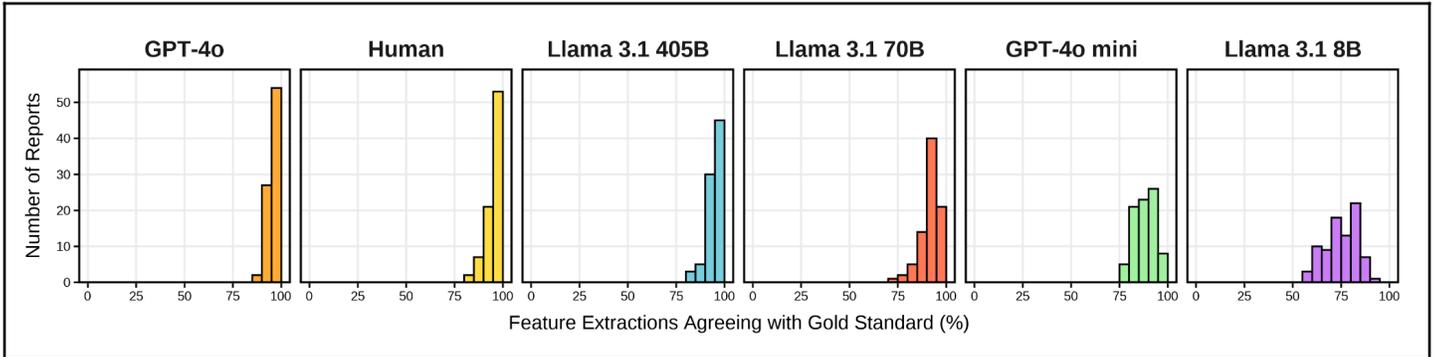

**Figure 5: Distribution of the percentage of extractions agreeing with the gold standard by report for each annotator.** The x-axis represents the percentage of fields accurately extracted within each report, while the y-axis indicates the number of reports that fall within each range of extraction accuracy.

The performance of all models across the reports followed roughly that of a binomial distribution, with no model performing exceptionally poorly on any individual report relative to its overall performance (Figure 5). The lowest-performing model, Llama 3.1 8B, aligned with the gold standard at a minimum of 57% on one of the 83 reports, while the highest-performing model, GPT-4o, achieved a minimum alignment of 88%.

## 4. Discussion

### 4.1 Extraction accuracy and costs

This section examines the performance-cost trade-offs of the models studied (Figure 6). Performance is measured by feature-value annotation accuracy across all features and reports, compared to the gold standard annotation. Costs are calculated as average per report in US dollars, based on OpenAI API rates for GPT-4o model family and Amazon Bedrock pricing for

Llama 3 models. Note that Llama 3 models can be self-hosted, potentially incurring no direct costs when deployed on user-owned computing resources.

The results from our evaluation demonstrate that certain LLMs can achieve human-level performance in extracting structured information from pathology reports. Specifically, both Llama 3.1 405B and GPT-4o demonstrated performance comparable to a trained human annotator, with 94.7% and 96.1% agreement with the gold standard, respectively.

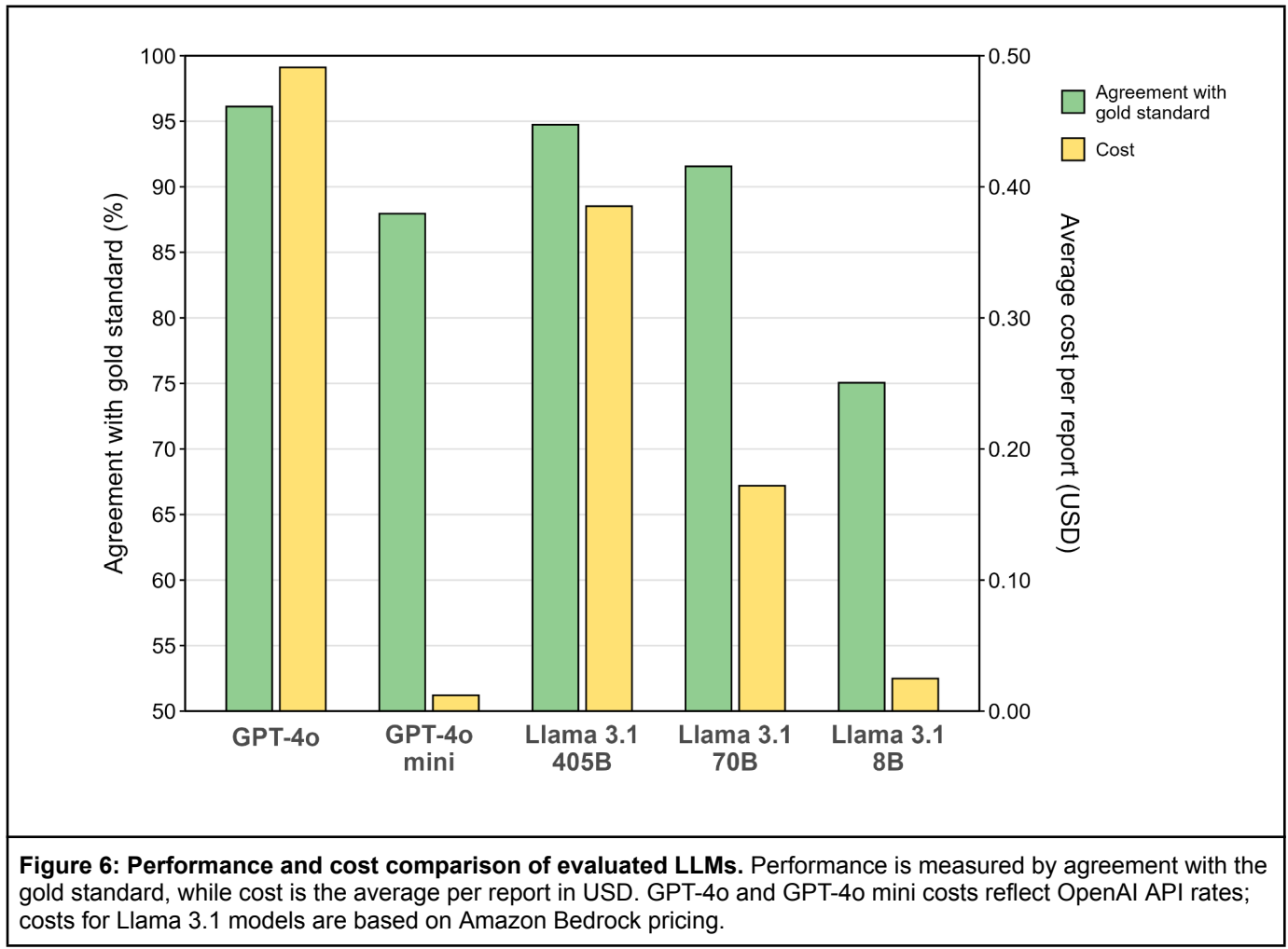

**Figure 6: Performance and cost comparison of evaluated LLMs.** Performance is measured by agreement with the gold standard, while cost is the average per report in USD. GPT-4o and GPT-4o mini costs reflect OpenAI API rates; costs for Llama 3.1 models are based on Amazon Bedrock pricing.

When comparing models within the GPT-4o family, GPT-4o significantly outperformed GPT-4o mini (96.1% vs 87.9%), though this came with a significantly higher cost per report ($0.44 vs $0.01). GPT-4o may be an attractive choice for applications prioritizing accuracy, while GPT-4o mini offers a viable alternative for applications where cost efficiency is key and a moderate error rate is acceptable. That said, the proprietary nature of both models and their reliance on OpenAI-hosted APIs may require specific data privacy considerations.

Among the Llama 3.1 models, performance increases with model size (8B: 75%, 70B: 91.6%, 405B: 94.7% agreement), as does cost per report ($0.02, $0.15, and $0.35 respectively), demonstrating a clear relationship between accuracy and computational expense.

Llama 3.1 405B matched the human annotator's performance without dropping below 80% accuracy on any report. On performance alone, Llama 3.1 405B ranks just behind GPT-4o, making it the best performing self-hostable model. However, its substantial computational requirements may limit local deployment options. For many users, accessing Llama 3.1 405B through managed services like Amazon Bedrock, Azure AI Studio, or Google Cloud Platform's VertexAI may be the most viable solution.

Considering its admirable performance and reasonable size for self-hosting, Llama 3.1 70B stands as a compelling alternative, balancing performance and cost-effectiveness. Self-hosting this model could eliminate both the need for data de-identification and direct costs from API use, making it a particularly attractive choice for practical implementation.

Among the self-hostable models assessed, Llama 3.1 8B stands out as the smallest in computational footprint. Its size allows for deployment entirely within the privacy of the user's web browser, as demonstrated by the WebLLM project. [27] However, its performance was notably inconsistent, with accuracy dropping to nearly 50% on some reports. While the performance of portable models currently lags behind their larger counterparts, their rapid recent development is noteworthy and they are expected to continue improving in performance. [8] The development of portable models merits continued observation and evaluation, as they may offer unique advantages in scenarios where data privacy requirements are restrictive and local processing is desirable.

Information extraction by a human requires prerequisite training, is limited by lower throughput, and is comparatively more expensive. In contrast, LLMs are available off-the-shelf as managed services, their throughput is only limited by the budget and hardware availability, and carry a lower cost per report.

It is important to note that, given the rapidly evolving technological landscape with model performance continually improving and costs rapidly declining, it is entirely possible that the recommendations based on these cost/performance analyses could change in the near future.

## 4.2 Challenges and limitations

We found that some disagreements observed across annotators were likely due to ambiguities in the original data dictionary. Conflicting information for the same feature was often found across different sections of the reports, such as "clinical details", "macroscopic report", "microscopic report", and "summary", without clear instructions on how to resolve these discrepancies. The *Whole Tumor Size* and *Closest Margin* features were noted to be particularly affected by these conflicting sections. Further complicating the interpretation of some features, the data dictionary did not provide clear guidance on whether these features referred to DCIS, invasive components, or whether one should be prioritized over the other. Disagreements in lymph node-related features were also likely affected by data dictionary ambiguities, particularly regarding the definitions of axillary vs. "other" nodes. Furthermore, some reports included open-ended ranges (e.g., <1mm, >10mm), but the data dictionary did not provide direction on how to handle these values. Improving the data dictionary could potentially improve the performance of the annotators by providing clearer guidance and reducing ambiguities.

OCR errors represent a potential concern when relying on automated methods for information extraction. In this study, OCR errors were evaluated only in cases where the human annotator and GPT-4o disagreed, with approximately 7% of these instances (23 out of 348) affected by OCR errors. In these cases, the gold standard was assigned to the human annotator's extraction. Since disagreements accounted for only a small fraction of the total extractions, and cases where the human annotator and GPT-4o agreed were presumably less affected by OCR errors, this 7% estimate likely overestimates the true OCR error rate. Additionally, because the gold standard was derived from OCR-processed plaintext data, undetected OCR errors, missed handwritten text, or information inadvertently removed during de-identification may have introduced a bias against the human annotator, who had access to the original, unprocessed information. As a result, the human annotator may have been incorrectly judged as making errors relative to the gold standard, even when the human had correctly extracted the information.

The gold standard dataset in this evaluation carried an inherent bias in favor of both GPT-4o and the human annotator, as their agreements were presumed correct without accounting for potential shared errors. This bias can be attributed to two sources: one, due to shared errors made by chance, and another due to errors resulting from underlying processes shared exclusively by GPT-4o and the human. However, given the fundamental differences between GPT-4o and human cognition, any observed interdependence likely resulted from external factors like data quality issues, which would affect all evaluated models similarly. Therefore, the primary remaining bias likely arose due to independent random chance alone.

## 4.3 AI systems in the foundation models era

We developed the Medical Report Information Extractor, an AI system driven by LLMs, which are a type of foundation model. Foundation models are generalist models adaptable to various specialized tasks, which inherit their benefits. Although a training dataset was used to refine our prompts, our approach using zero-shot prompting does not necessitate a training set. Zero-shot prompting is a prompting strategy where no examples are explicitly included within the prompts. The LLMs matching expert annotators' performance demonstrates its inherent and strong generalization capability to our task. LLMs' linguistic capabilities democratize the technology to non-programmers, like pathologists, to modify the AI system behavior using natural language, without the need for programming or data science expertise. Developers benefit from reduced development time since no data collection or model training is required. Such deployed AI systems undergo automatic continuous development as newer and better

LLMs become available. Additionally, commoditization of LLMs through managed services APIs have made these models widely accessible across industries and applications.

While the aforementioned benefits of these generalist foundation models are inherited by all specialized applications that adapt them, so are the risks. [5] As these generalist LLMs are developed without specific use cases in mind, newer versions may inadvertently prioritize certain use cases over others. This highlights the need for extensive validation, versioning, and application-specific benchmarking. Applications like Medical Report Information Extractor must be validated on larger and diverse datasets to help identify and preempt their failure modes. They should use API endpoints linked to specific LLM versions to prevent unexpected performance changes over time. Maintaining private, task-specific benchmarks is crucial to ensure safe upgrades to newer LLMs without unexpected failures. We addressed this by curating a gold standard dataset from our evaluation pipeline to test newer models before upgrading the application with them.

Self-hostable LLMs provide enhanced data privacy compliance, eliminate the need for data de-identification, and reduce costs compared to proprietary solutions. This is particularly beneficial for large-scale applications. Consequently, validation studies should assess the performance of self-hostable alternatives alongside proprietary models.

The Medical Report Information Extractor web application was designed with the versatility of foundation models in mind. Its behavior is configurable through three external human-readable files, allowing adaptation to various specialized information extraction tasks. The application demonstrates domain-agnostic flexibility, functioning independently of input document types (e.g., radiology reports, clinical notes, news articles), underlying LLMs, specified extraction features and formatting, and approach to data standardization. This design philosophy promotes broad reusability and ease of customization across diverse information extraction tasks.

In our application, both the extracted data (JSON/JSON-LD) and the data dictionary (JSON Schema) use extensions of the JSON format. As a result, both formats remain non-proprietary, and both human- and machine-readable. JSON allows us to store both the data and metadata using document-based databases and distribute it through standard RESTful APIs, supporting interoperability and integration with existing systems and workflows.

## 4.4 Future work

An important direction for future research is to evaluate the generalizability of the structured information extraction pipeline using larger datasets across a broader range of medical records, including radiology reports and clinical notes, as well as expanding to non-medical domains. It will also be essential to continue testing newer LLMs as they are released, using our gold standard testing dataset.

In our assessment, we used simple zero-shot prompting. While this approach provided a baseline for performance, more advanced prompting techniques, such as MedPrompt, [4] have shown potential to significantly improve performance over simpler prompting strategies. To achieve further performance gains, a promising direction for future work is to explore the use of fine-tuned models for specific information extraction tasks. While fine-tuning does require task-specific data, it remains considerably more sample-efficient than training conventional NLP models from scratch, as was common before the advent of foundation models.

MedGemini, a proprietary model fine-tuned on extensive medical data, has been shown to outperform general-purpose models on certain tasks. [31,32] Future work could evaluate such domain-specific models on specific information extraction tasks to determine if they outperform the general-purpose LLMs assessed in our study. Similarly, even smaller fine-tuned models have demonstrated the ability to outperform larger general-purpose models in some situations, [33] suggesting that fine-tuning the Llama 3.1 70B model on task-specific data could be a promising avenue for further exploration. Alternatively, models trained from scratch for general-purpose structured information extraction, such as the portable NuExtract model, have shown performance on par with GPT-4, though it has a limited context length. [34]

Exploring large multimodal models (LMMs) presents an opportunity to streamline the structured information extraction process by potentially eliminating the need for separate optical character recognition (OCR) and layout alignment steps. However, current LMMs still fall short of the performance achieved by models trained specifically for text recognition. [35]

Evaluating the ability of LLMs to handle reports with multiple specimen descriptions is another important area for future study. The current research focused solely on single-specimen reports; assessing whether LLMs can accurately assign extracted features to the correct specimen in multi-specimen reports will determine their abilities in more complex documents. Extending the evaluation to non-English reports is also crucial for assessing the multilingual information extraction capabilities of LLMs. Previous trials on German radiology reports have shown promise. [36]

In this study, a human prompt engineer manually performed iterative prompt development. Future studies could explore automating this step through LLMs seeded only with the data dictionary. The predictive performance in the training set, along with examples of errors in extraction, can be used to prompt the LLM towards iteratively developing prompts that improve predictive performance.

There has been a recent emergence of a class of language models that utilize extended runtime computation to simulate complex reasoning through structured deliberation. [37] Examples of such models include OpenAI's o1 and o3, Google's Gemini 2.0 Flash Thinking, and DeepSeek-R1. Their enhanced reasoning capabilities come with significantly higher latency and costs. Future work may explore optimal strategies to employ these models for structured information extraction.

Another potential direction to improve the portability of the application is to evaluate LLMs that run locally on user devices. WebLLM, an in-browser inference engine, offers a portable, cross-platform, and secure solution through a sandboxed browser environment. [27] However, running models on-device may introduce hardware limitations that need to be considered.

# 5. Conclusion

This investigation compared the performance of state-of-the-art LLMs to a trained human annotator in accurately extracting structured information from real-world pathology reports. The LLMs were adapted for the task using zero-shot prompting, without additional training. We assessed 111 histopathology reports from the UK BCN Generations Study, extracting 51 features of interest specified in the study's data dictionary. Additionally, the data dictionary was standardized to SNOMED CT terminology using a separate LLM-driven technique, enhancing its interoperability for collaborative research.

Using a gold standard extraction dataset that we developed, our evaluation found that both the self-hostable Llama 3.1 405B model and the proprietary GPT-4o model achieved performance comparable to that of a trained human annotator in terms of extraction accuracy. For resource-constrained settings, the smaller-sized Llama 3.1 70B model offers a compelling alternative, balancing competitive performance with the improved feasibility of self-hosting, which eases data privacy concerns by eliminating the need to share confidential information externally. Our findings also highlight the importance of a comprehensive, well-defined data dictionary in ensuring accurate information extraction and minimizing manual prompt refinement.

To employ LLMs for structured information extraction, we developed and open-sourced the Medical Report Information Extractor web application. The application and instructions for its use are available at: https://jeyabbalas.github.io/medical-report-information-extractor/. The source code is available at: https://github.com/jeyabbalas/medical-report-information-extractor. The application's behavior is easily configurable by editing a set of human-readable files, allowing adaptation to a wide range of information extraction tasks. This design democratizes the technology for domain-experts, such as pathologists, enabling them to build and modify customized information extraction applications using only natural language interfaces, without the need for programming expertise. We anticipate that this design will motivate its reuse in similar applications, enhancing data accessibility, interoperability, and analytics in clinical research by increasing the availability of standardized, structured data.

# Funding support

The study is funded by Intramural Funds of the US National Cancer Institute (Maryland, USA), Breast Cancer Now (London, UK), and The Institute of Cancer Research (London, UK).

# Declaration of competing interest

No competing interests.

# Acknowledgments

We acknowledge the work of Ms. Jane Lebihan in collecting pathology reports and performing manual data abstractions for the Breast Cancer Now Generations Study. We also acknowledge the participants, study staff, and health care staff for their contribution to the study.

# Supplement

**Table S1: Definitions of acronyms used in pathology report feature names.**

| Acronym | Definition |
| --- | --- |
| CIS | Carcinoma In Situ |
| DCIS | Ductal Carcinoma In Situ |
| ER | Estrogen Receptor |
| FISH | Fluorescence In Situ Hybridization |
| GP | Growth Pattern |
| HER2 | Human Epidermal Growth Factor Receptor 2 |
| IHC | Immunohistochemistry |
| LCIS | Lobular Carcinoma In Situ |

**Table S2. Accuracy of each annotator in extracting information by different pathology report features.** For each feature, the table shows the accuracy percentages of the human annotator, GPT-4o, GPT-4o mini, and the Llama 3.1 models (405B, 70B, and 8B) in terms of agreement with the gold standard. The "Mention" column represents the percentage of reports in which each feature was mentioned. See Table S1 for definitions of feature-related acronyms.

| Feature | Annotators | | | | | | Mention |
| --- | --- | --- | --- | --- | --- | --- | --- |
| | Human | GPT-4o | GPT-4o mini | Llama 3.1 405B | Llama 3.1 70B | Llama 3.1 8B | |

| Patient and Specimen Information | | | | | | | |
| --- | --- | --- | --- | --- | --- | --- | --- |
| Side | 97.59% | **100.00%** | 98.80% | **100.00%** | **100.00%** | 98.80% | 100.00% |
| Screen Detected | **97.59%** | 95.18% | 92.77% | **97.59%** | 91.57% | 65.06% | 16.87% |
| Specimen Type | 97.59% | **98.80%** | 95.18% | 96.39% | 95.18% | 91.57% | 100.00% |
| Specimen Weight | **97.56%** | 95.12% | 91.46% | 92.68% | 90.24% | 79.27% | 79.27% |
| Post-Neoadjuvant Chemo | **98.80%** | 97.59% | 97.59% | **98.80%** | 96.39% | 97.59% | 2.41% |
| **In Situ Components** | | | | | | | |
| CIS | 96.39% | 98.80% | 63.86% | **100.00%** | 97.59% | 63.86% | 96.39% |
| DCIS | 96.39% | **100.00%** | 97.59% | 97.59% | 97.59% | 84.34% | 95.18% |
| LCIS | 90.36% | 89.16% | 86.75% | **93.98%** | 90.36% | 43.37% | 57.83% |
| Pleomorphic LCIS | **100.00%** | 96.39% | 97.59% | 98.80% | 97.59% | 31.33% | 0.00% |
| Paget's Disease | **98.80%** | 96.39% | 80.72% | 85.54% | 74.70% | 51.81% | 50.60% |
| Microinvasion | **95.18%** | 93.98% | 86.75% | 92.77% | 91.57% | 24.10% | 19.28% |
| **DCIS Characteristics** | | | | | | | |
| Solid DCIS GP | 95.18% | 98.80% | 86.75% | 98.80% | **100.00%** | 89.16% | 48.19% |
| Cribriform DCIS GP | 96.39% | 98.80% | 93.98% | 98.80% | **100.00%** | 93.98% | 30.12% |
| Micropapillary DCIS GP | **98.80%** | **98.80%** | 96.39% | 97.59% | **98.80%** | 95.18% | 8.43% |
| Papillary DCIS GP | **100.00%** | **100.00%** | 98.80% | 98.80% | 98.80% | **100.00%** | 2.41% |
| Apocrine DCIS GP | **100.00%** | **100.00%** | 97.59% | 97.59% | **100.00%** | 97.59% | 0.00% |
| Flat DCIS GP | **100.00%** | **100.00%** | **100.00%** | **100.00%** | **100.00%** | 98.80% | 0.00% |
| Comedo DCIS GP | 91.57% | **98.80%** | 92.77% | **98.80%** | 91.57% | 73.49% | 14.46% |
| Clinging DCIS GP | **100.00%** | **100.00%** | 98.80% | **100.00%** | **100.00%** | **100.00%** | 1.20% |
| Comedonecrosis Extent | **98.80%** | 75.90% | 73.49% | **98.80%** | 95.18% | 80.72% | 32.53% |
| **Size, Extent, and Margins** | | | | | | | |
| Tumor Extent | 93.98% | **98.80%** | **98.80%** | 97.59% | 92.77% | 81.93% | 56.63% |
| Whole Tumor Size | **86.75%** | 85.54% | 80.72% | 77.11% | 79.52% | 78.31% | 77.11% |
| DCIS Size | 93.98% | **100.00%** | 95.18% | 96.39% | 96.39% | 83.13% | 14.46% |
| Invasive Tumor Size | 93.98% | **100.00%** | 97.59% | **100.00%** | 98.80% | 92.77% | 86.75% |
| Closest Margin | **85.54%** | 81.93% | 73.49% | 75.90% | 72.29% | 45.78% | 80.72% |
| Excision Margin | 92.77% | **93.98%** | 90.36% | 90.36% | 89.16% | 69.88% | 92.77% |
| **Invasive Disease Characteristics** | | | | | | | |
| Invasive Carcinoma | **100.00%** | **100.00%** | **100.00%** | 98.80% | 98.80% | 97.59% | 100.00% |
| Invasive Carcinoma Type | 87.95% | **98.80%** | 96.39% | 93.98% | 91.57% | 69.88% | 90.36% |
| Invasive Carcinoma Subtype | 91.57% | **100.00%** | 96.39% | 93.98% | 93.98% | 89.16% | 15.66% |
| **Tumor Grading and Scoring** | | | | | | | |
| Invasive Grade | **98.80%** | 97.59% | **98.80%** | 96.39% | 96.39% | 96.39% | 90.36% |
| Tubules Score | **97.59%** | 90.36% | 68.67% | 96.39% | 86.75% | 72.29% | 72.29% |
| Nuclear Pleomorphism Score | **97.59%** | 96.39% | 68.67% | 95.18% | 83.13% | 68.67% | 72.29% |
| Mitotic Activity Score | 95.18% | 95.18% | 72.29% | **96.39%** | 83.13% | 73.49% | 73.49% |
| DCIS Grade | 92.77% | **98.80%** | 93.98% | 95.18% | **98.80%** | 80.72% | 71.08% |
| **Staging Information** | | | | | | | |
| T Stage | **96.39%** | 92.77% | 62.65% | 71.08% | 72.29% | 55.42% | 72.29% |
| N Stage | **97.59%** | 93.98% | 68.67% | 86.75% | 60.24% | 72.29% | 72.29% |
| M Stage | **100.00%** | 89.16% | 37.35% | 92.77% | 67.47% | 31.33% | 30.12% |
| **Lymphovascular Involvement** | | | | | | | |
| Vascular Invasion | 96.39% | 93.98% | 91.57% | **100.00%** | 98.80% | 91.57% | 89.16% |
| Axillary Nodes Present | 89.16% | **97.59%** | 78.31% | 95.18% | 92.77% | 86.75% | 93.98% |
| Axillary Nodes Total | 90.36% | **95.18%** | 79.52% | 89.16% | 86.75% | 34.94% | 78.31% |
| Axillary Nodes Positive | 90.36% | **100.00%** | 86.75% | 93.98% | 93.98% | 77.11% | 79.52% |
| Other Nodes Present | **92.77%** | 89.16% | 80.72% | 90.36% | 79.52% | 38.55% | 27.71% |
| Other Nodes Total | 95.18% | **98.80%** | 96.39% | 93.98% | 95.18% | 61.45% | 9.64% |
| Other Nodes Positive | 93.98% | **100.00%** | 95.18% | 96.39% | 95.18% | 65.06% | 10.84% |
| **Receptor and Biomarker Status** | | | | | | | |
| ER Score Type | **97.59%** | **97.59%** | 93.98% | **97.59%** | **97.59%** | 90.36% | 90.36% |
| ER Status | **98.80%** | **98.80%** | **98.80%** | **98.80%** | **98.80%** | 92.77% | 95.18% |
| PR Score Type | 93.98% | **98.80%** | 91.57% | **98.80%** | 97.59% | 78.31% | 77.11% |
| PR Status | 96.39% | 96.39% | 93.98% | **97.59%** | 95.18% | 75.90% | 78.31% |
| HER2 IHC Score | 93.98% | **96.39%** | 91.57% | 90.36% | 89.16% | 74.70% | 65.06% |
| HER2 FISH | 87.95% | **95.18%** | 79.52% | 91.57% | 85.54% | 72.29% | 31.33% |
| Ki-67 | **100.00%** | 98.80% | **100.00%** | **100.00%** | 95.18% | 68.67% | 16.87% |